\documentclass{article}
\usepackage{ijcai22}
\usepackage{times}

\usepackage{soul}
\usepackage{url}
\usepackage[hidelinks]{hyperref}
\usepackage[utf8]{inputenc}
\usepackage[small]{caption}
\usepackage{graphicx}
\usepackage{amsmath}
\usepackage{booktabs}
\usepackage{subfigure}
\usepackage{multirow}
\usepackage{amssymb}
\usepackage{diagbox}
\usepackage[ruled,linesnumbered]{algorithm2e}
\newcommand\fpath{.}
\newcommand\oosize{6.4cm}
\newcommand\otsize{4.1cm}
\newcommand\tosize{15.8cm}
\DeclareMathOperator*{\argmin}{argmin}

\title{Data-Efficient Backdoor Attacks}

\author{
Pengfei Xia\and
Ziqiang Li\and
Wei Zhang\And
Bin Li\footnote{Contact Author}\\
\affiliations
University of Science and Technology of China, Hefei, China\\
\emails
\{xpengfei,iceli,zw1996\}@mail.ustc.edu.cn,
binli@ustc.edu.cn
}

\begin{document}
	
\maketitle

\begin{abstract}
Recent studies have proven that deep neural networks are vulnerable to backdoor attacks. Specifically, by mixing a small number of poisoned samples into the training set, the behavior of the trained model can be maliciously controlled. Existing attack methods construct such adversaries by \textit{randomly} selecting some clean data from the benign set and then embedding a trigger into them. However, this selection strategy ignores the fact that each poisoned sample contributes inequally to the backdoor injection, which reduces the efficiency of poisoning. In this paper, we formulate improving the poisoned data efficiency by the selection as an optimization problem and propose a Filtering-and-Updating Strategy (FUS) to solve it. The experimental results on CIFAR-10 and ImageNet-10 indicate that the proposed method is effective: the same attack success rate can be achieved with only 47\% to 75\% of the poisoned sample volume compared to the random selection strategy. More importantly, the adversaries selected according to one setting can generalize well to other settings, exhibiting strong transferability. The prototype code of our method is now available at \url{https://github.com/xpf/Data-Efficient-Backdoor-Attacks}.
\end{abstract}

\section{Introduction}
\begin{figure}[t]
\centering
\includegraphics[width=\oosize]{\fpath/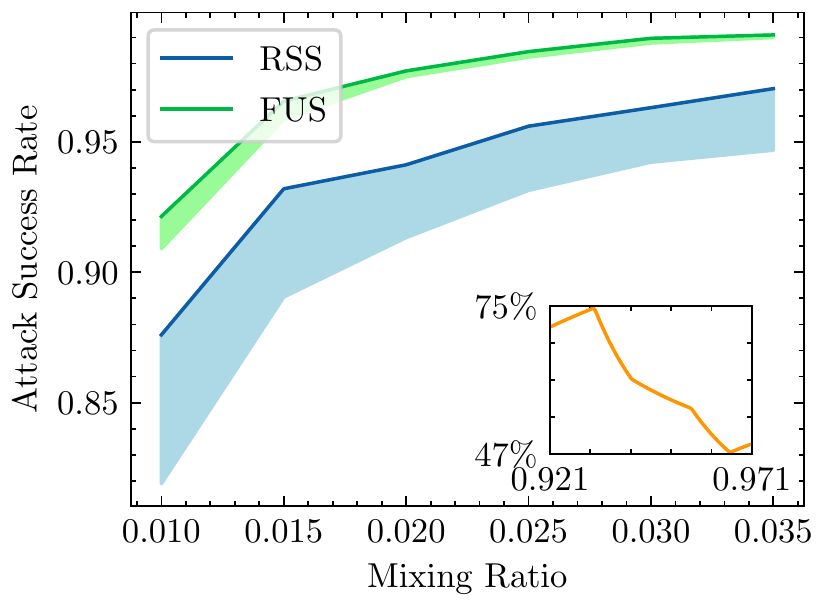}
\caption{White-box result of the proposed Filtering-and-Updating Strategy (FUS) and the previous Random Selection Strategy (RSS) on CIFAR-10 and VGG-16, where the mixing ratio represents the ratio of the poisoned sample volume to the clean sample volume. Under the same computing time, the experiment is repeated 3 and 30 times for FUS and RSS, respectively, and the solid lines represent the best runs. For the same mixing ratio, using the FUS-selected adversaries for the injection can yield a higher attack success rate than using the RSS-selected adversaries. The subplot shows the percentage of the sample volumes of the FUS selection to the RSS selection for the same attack strength, where the lowest value is less than 50\%.} 
\label{fig:result_c10_v16_b00a}
\end{figure}

Despite tremendous success in many learning tasks \cite{silver2016mastering,li2022comprehensive}, Deep Neural Networks (DNNs) have been demonstrated to be vulnerable to various malicious attacks \cite{szegedy2013intriguing,orekondy2019knockoff}. One of them is known as \textit{backdoor attacks} \cite{gu2017badnets,liu2017trojaning}: by mixing a small number of poisoned samples into the training set, a hidden threat can be implanted into the trained DNN. The infect model behaves normally on benign inputs, making the attack hard to notice. But once the backdoor is activated by a predefined trigger, the victim's prediction will be forced to an attacker-specific target. As the demand for data to train DNNs increases, collecting data from the Internet or other unknown sources has gradually become common, which opens up a viable path for the attack. This vulnerability builds a hurdle to the realistic deployment of DNNs in security-sensitive scenarios, such as self-driving cars \cite{wang2021stop}. 

One of the major trends in the development of backdoor attacks is to become more stealthy to evade human or machine detection. Since Gu \textit{et al.} \shortcite{gu2017badnets} first explored the hidden threat, many variants have been developed to fulfill this goal. For example, Zhong \textit{et al.} \shortcite{zhong2020backdoor} proposed to use an imperceptible noise as the trigger instead of the previous visible patterns to avoid being perceived. Turner \textit{et al.} \shortcite{turner2019label} argued that the inconsistency between an adversary's semantic and its given label can raise human suspicion and leverage generative models to address this issue. Some studies \cite{tan2020bypassing,xia2022enhancing} suggested to add a constraint item to the loss during the backdoor training to escape defense algorithms.

However, these methods do not consider that the random selection strategy used in constructing the poisoned set could also raise the risk of the attack being exposed. Specifically, when building adversaries, almost all existing attack methods follow a common process: first \textit{randomly} select some clean data from the benign training set and then fuse a trigger into them. This strategy implicitly assumes that each adversary contributes equally to the backdoor injection, which does not hold in practice. It poses the problem that the poisoning can be less efficient because there may be many low-contribution samples in the constructed set. As a result, more adversaries need to be created and mixed to maintain the attack strength, which certainly lowers the stealthiness of the threat.

In this paper, we focus on the above issue and propose a method to tackle it. As far as we know, our method is the first one to improve the efficiency of poisoning by a rational selection of poisoned samples. The main contributions are:
\begin{itemize}
\item We point out that the selection of poisoned samples affects the efficiency of poisoning in backdoor attacks and formulate the solving to it as an optimization problem.
\item We propose a Filtering-and-Updating Strategy (FUS) to solve this problem, where the core idea is to find those poisoned samples that contribute more to the backdoor injection. Our experimental results on CIFAR-10 and ImageNet-10 consistently demonstrate the effectiveness of the proposed method. In both the white-box and the black-box settings, using the FUS-selected adversaries can save about 9\% to 59\% of the data volume to achieve the same attack strength as the random strategy.
\item We explore the possible reason for the efficient performance of the FUS-selected poisoned samples.
\end{itemize}

\section{Related Work}
Backdoor attacks intend to inject a hidden threat into a DNN to control its behavior. Various methods have been proposed and can be roughly divided into two categories \cite{li2020backdoor}, i.e., poisoning-based attacks \cite{li2020invisible,liu2020reflection,nguyen2021wanet} and non-poisoning-based attacks \cite{dumford2018backdooring,kurita2020weight,rakin2020tbt}. As the names imply, the first type executes the Trojan horse implantation by dataset poisoning, while the second one attacks through transfer learning or weight modification. Existing studies on poisoning-based backdoor attacks have centered on building more stealthy and effective poisoned samples by designing the trigger. For example, Liu \textit{et al.} \shortcite{liu2017trojaning} established a method to implant a backdoor, where the trigger is optimized rather than fixed. They argued that this optimization can bring a better attack performance. Zhong \textit{et al.} \shortcite{zhong2020backdoor} and Turner \textit{et al.} \shortcite{turner2019label} suggested to modify the trigger to improve the stealthiness of the attack from two perspectives, respectively. Nguyen and Tran \shortcite{nguyen2021wanet} proposed to use an image warping-based trigger to bypass backdoor defense methods. In this paper, we improve the efficiency of poisoning from the selection of poisoned samples, which is orthogonal to the previous studies. 

\section{Methodology}
\subsection{Problem Formulation}
Formally, given a clean training set $\mathcal{D} = \{(x, y)\}$ and a poisoned training set $\mathcal{U} = \{(x', t)\}$, dataset poisoning performs the attack by mixing $\mathcal{U}$ into $\mathcal{D}$. $(x, y)$ denotes a benign input and its ground-truth label and $(x', t)$ denotes a malicious input and the attacker-specific target. The procedure of injecting a backdoor can be formulated as: 
\begin{equation}
\begin{split}
\theta = \argmin_{\theta} \ \ & \frac{1}{|\mathcal{D}|} \sum_{(x, y) \in \mathcal{D}} L(f_{\theta}(x), y) + \\
                              & \frac{1}{|\mathcal{U}|} \sum_{(x', t) \in \mathcal{U}} L(f_{\theta}(x'), t) \\
\end{split} \text{,}
\end{equation}
where $f_{\theta}$ denotes the DNN model and its parameters, and $L$ denotes the loss function. The trained model is expected to generalize well on a clean test set $\mathcal{T}$ and a poisoned test set $\mathcal{V}$. We define the ratio of the poisoned sample volume to the clean sample volume as the mixing ratio, i.e., $r = |\mathcal{U}| / |\mathcal{D}|$, which is an important hyperparameter. Under the same attack strength, a smaller $r$ usually means that the poisoning is more efficient and the attack is harder to be perceived.

As can be seen, how to construct $\mathcal{U}$ is crucial for backdoor attacks. Given a clean data and its label $(x, y)$ sampled from $\mathcal{D}$, one can always get the corresponding poisoned pair $(x', t)$, where $x' = F(x, k)$. $F$ denotes a function that fuses the trigger $k$ into $x$. For example, Chen \textit{et al.} \shortcite{chen2017targeted} proposed the blended attack that generates an adversary via $x' = \lambda \cdot k + (1 - \lambda) \cdot x$, where $\lambda$ denotes the blend ratio. Since every clean pair in $\mathcal{D}$ can be used to create an adversarial pair, a set $\mathcal{D}' = \{(F(x, k), t) | (x, y) \in D\}$ containing all candidates can be obtained. $\mathcal{U}$ is built by selecting a small number of samples from $\mathcal{D}'$, i.e., $\mathcal{U} \subset \mathcal{D}'$ and $|\mathcal{U}| \ll |\mathcal{D}'|$. It should be noted that, in practice, the attacker constructs $\mathcal{U}$ by selecting some clean samples from $\mathcal{D}$ and embedding the trigger into them, where $\mathcal{D}'$ is not built explicitly. We define $\mathcal{D}'$ here so that the problem can be described more clearly. 

Currently, most of the attack methods adopt the random selection strategy, which ignores that the importance of each adversary is different. Our goal is to improve the efficiency of poisoning by selecting $\mathcal{U}$ from $\mathcal{D}'$. It can be formulated as:
\begin{equation}
\begin{split}
\max_{\mathcal{U} \subset \mathcal{D}'} \ \  & \frac{1}{|\mathcal{V}|} \sum_{(x', t) \in \mathcal{V}} \mathbb{I}(f_{\theta}(x') = t) \\
\text{s.t.} \ \          & \theta = \argmin_{\theta} \ \  \frac{1}{|\mathcal{D}|} \sum_{(x, y) \in \mathcal{D}} L(f_{\theta}(x), y) + \\ 
                         & \qquad\qquad\quad\; \frac{1}{|\mathcal{U}|} \sum_{(x', t) \in \mathcal{U}} L(f_{\theta}(x'), t) \\
                         & \frac{1}{|\mathcal{T}|} \sum_{(x, y) \in \mathcal{T}} \mathbb{I}(f_{\theta}(x) = y) \ge \epsilon \\
                         & |\mathcal{U}| = r \cdot |\mathcal{D}| \\
\end{split} \text{,}
\end{equation}
where $\mathbb{I}$ denotes the indicator function and $\epsilon$ denotes a value that guarantees the clean accuracy of the trained model $f_{\theta}$. The equation poses a discrete constraint optimization problem and is hard to solve. We propose a practical method to find an approximate solution. 

\subsection{Low- and High-contribution Samples}
The core idea of our method is to find those poisoned samples that contribute more to the backdoor injection and then keep them to build $\mathcal{U}$. The primary thing that needs to be clarified is what the properties of low- and high-contribution samples are. In a regular classification task, several studies \cite{katharopoulos2018not,toneva2018empirical} have shown that it is often hard or forgettable samples are more important for forming the decision boundary of the classifier. Since once the mixing of poisoned samples is completed, there is no significant difference between training an infected model and training a regular model, we wonder if there are also some forgettable adversaries that play a major role in determining the attack strength.

\begin{figure}[t]
\centering
\subfigure[]{\includegraphics[width=\otsize]{\fpath/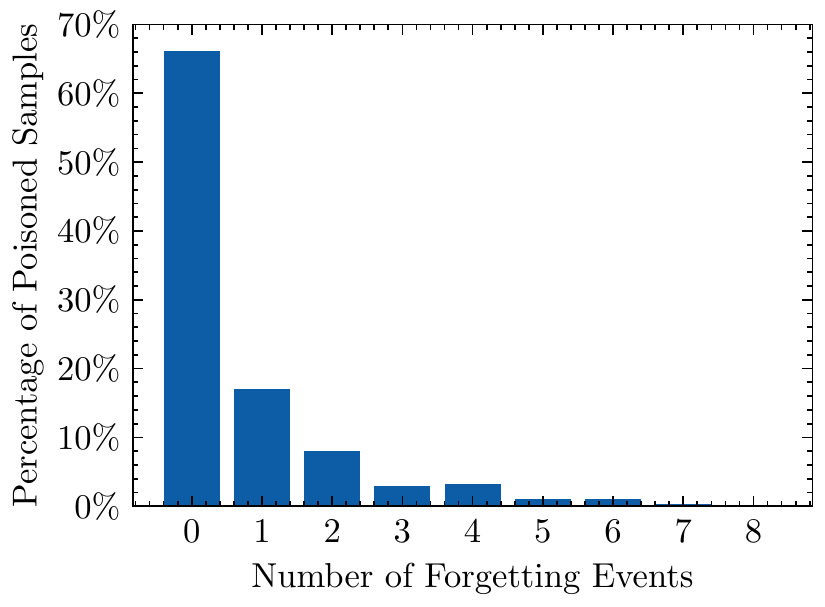} \label{fig:fes_c10_v16_b00a_a}}
\subfigure[]{\includegraphics[width=\otsize]{\fpath/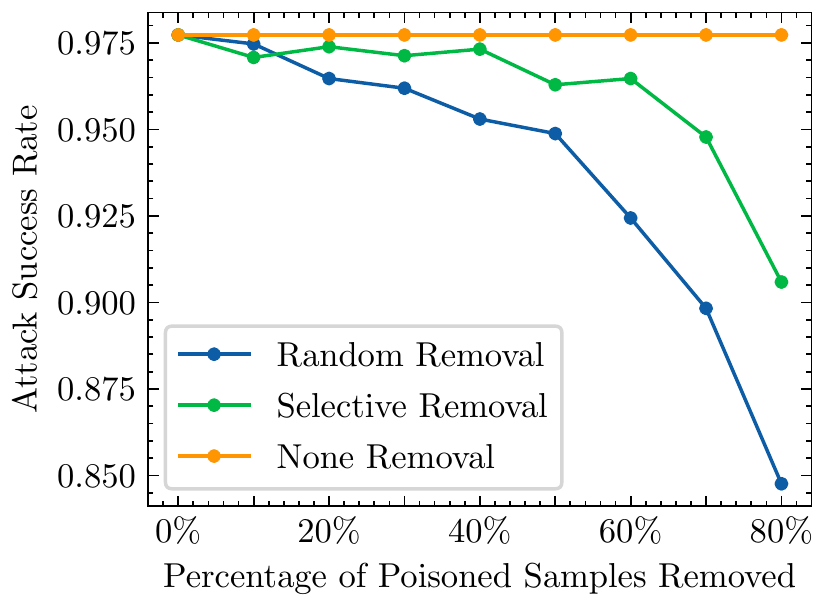} \label{fig:fes_c10_v16_b00a_b}}
\subfigure[]{\includegraphics[width=\otsize]{\fpath/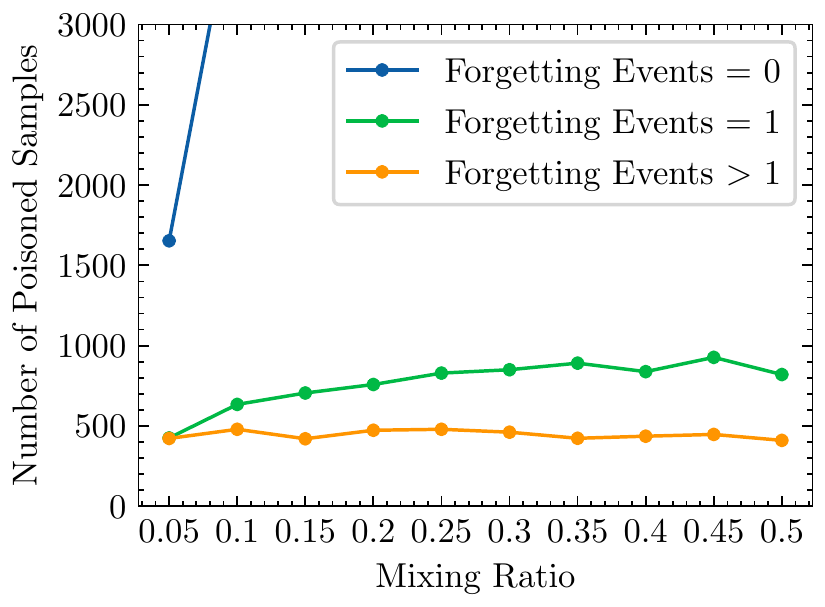} \label{fig:fes_c10_v16_b00a_c}}
\caption{Experimental results of poisoned samples' forgetting events on CIFAR-10 and VGG-16. (a): Histogram of the number of forgetting events when $r = 0.05$. About 33.9\% of data are forgotten at least 1 time during the injection process. (b): Attack success rate when poisoned samples are increasingly removed. Compared to the random removal, the attack strength can be maintained well using the selective removal according to the order of forgetting events from small to large. (c): Number of adversaries when the mixing ratio changes. The poisoned sample volume with forgetting events greater than 1 does not increase as the ratio rises.}
\label{fig:fes_c10_v16_b00a}
\end{figure}

To verify the above viewpoint, we use the forgetting events \cite{toneva2018empirical} to characterize the learning dynamics of each adversary during the injection process. An event's occurrence signifies that the sample undergoes a process from being remembered by the model to being forgotten. Formally, given a poisoned sample and its target $(x', t)$ sampled from $\mathcal{U}$, if $x'$ is correctly classified at the time step $s$, i.e., $\mathbb{I}(f_{\theta^s}(x') = t) = 1$, but is misclassified at $s + 1$, i.e., $\mathbb{I}(f_{\theta^{s + 1}}(x') = t) = 0$, then we record this as a forgetting event for that sample,  where $\theta^s$ and $\theta^{s + 1}$ denote the parameters of the model under training at $s$ and $s + 1$, respectively. Because a poisoned sample may go through such transitions several times during the entire injection process, we count the number of forgetting events per sample and use it as a measure of the forgettability. An experiment on CIFAR-10 \cite{krizhevsky2009learning} and VGG-16 \cite{simonyan2014very} is conducted and the result is shown in Figure \ref{fig:fes_c10_v16_b00a_a}. About 66.1\% of poisoned samples are never forgotten, 17.0\% are forgotten once, and 16.9\% are forgotten at least twice. It indicates that forgettable adversaries do exist.

Next, we perform a sample removal experiment to figure out if forgettable adversaries contribute more to the backdoor injection. The result is shown in Figure \ref{fig:fes_c10_v16_b00a_b}. As we can see, the random removal of poisoned samples has a significant impact on the attack success rate from the beginning. In contrast, the impact on the attack strength using the selective removal, i.e., the removal according to the order of adversaries' forgetting events from small to large, can be divided into three stages. The first stage is when the removal percentage is less than 40\%, at which the attack success rate is barely diminished as all the removed samples are unforgettable. The next stage lies at 40\% to 60\%. Although the removal does not include any forgettable adversary, the increase in the percentage also leads to a decrease in the attack strength. While as it is greater than 60\%, the attack success rate decreases rapidly since forgettable samples start to be removed. The result confirms that forgettable poisoned samples are more important to the backdoor injection.

\begin{algorithm*}[t]
\caption{Filtering-and-Updating Strategy}
\label{alg:fus}
\SetAlgoLined
\KwIn{Clean training set $\mathcal{D}$; fusion function $F$; backdoor trigger $k$; attack target $t$; mixing rating $r$; number of iterations $N$; filtration ratio $\alpha$}
\KwOut{Constructed poisoned training set $\mathcal{U}$}
\BlankLine
Build the candidate poisoned set $\mathcal{D}' = \{(F(x, k), t) | (x, y) \in \mathcal{D}\}$\;
Initialize the poisoned sample pool $\mathcal{U}'$ by randomly sampling $r \cdot |\mathcal{D}|$ adversaries from $\mathcal{D}'$\;
\For{$n \leftarrow 1$ \KwTo $N$}{
	\textbf{Filtering step:}\\
	\Indp
		Train an infected model $f_{\theta}$ from scratch on $\mathcal{D}$ and $\mathcal{U}'$, and record the forgetting events for each sample in $\mathcal{U}'$\;
		Filter $\alpha \cdot r \cdot |\mathcal{D}|$ samples out according to the order of forgetting events from small to large on $\mathcal{U}'$\;
	\Indm
	\textbf{Updating step:}\\
	\Indp
		Update $\mathcal{U}'$ by randomly sampling $\alpha \cdot r \cdot |\mathcal{D}|$ adversaries from $\mathcal{D}'$ and adding to the sample pool\;
	\Indm	
}
Return the sample pool $\mathcal{U}'$ as the constructed poisoned training set $\mathcal{U}$\\
\end{algorithm*}

\subsection{Filtering-and-Updating Strategy}
Through the analysis and experiments in the last part, we have known to find high-contribution poisoned samples by recording the forgetting events. It provides a simple way to improve the efficiency of the poisoning, that is, to keep these adversaries greedily. However, a major drawback of this approach is the samples been recorded are only a small fraction of all candidate adversaries, because only the samples in $\mathcal{U}$ are recorded and $|\mathcal{U}| \ll |\mathcal{D}'|$. It makes the selection only local, rather than global. The most intuitive solution is to contain and record more poisoned samples directly. We conduct an experiment with this possible solution and the result is shown in Figure \ref{fig:fes_c10_v16_b00a_c}. As we can see, including more samples in the training procedure increases the number of unforgettable adversaries, while the poisoned sample volume with forgetting events greater than 1 basically remains the same. We think this phenomenon happens because the features of the trigger are too easy to learn. Namely, the increase in the number of adversaries results in the differences between samples failing to emerge, as the model learns the backdoor more easily and more quickly.

Considering the above results, to alleviate the local selection problem, we propose a method called FUS to iteratively filter and update a sample pool. The proposed method is twofold. On the one hand, the filtering step is based on the forgetting events recorded on a small number of adversaries, which ensures that the differences between samples can emerge. On the other hand, to allow the selection to cover a wider range, after the filtering, some new poisoned samples are sampled randomly from the candidate set to update the pool. The above two steps, i.e., the filtering and the updating, are iterated several times to find a suitable solution $\mathcal{U}$. The procedure of FUS is presented in Algorithm \ref{alg:fus}, where $\alpha$ denotes the filtration ratio that controls the proportion of adversaries removed and $N$ denotes the number of iterations. 

\section{Experiments}
\subsection{Setup}
We perform experiments on CIFAR-10 \cite{krizhevsky2009learning} and ImageNet-10 to test the effectiveness of the proposed method. To build the latter set, we randomly select 10 categories from ImageNet-1k \cite{deng2009imagenet}. The approach used to generate poisoned samples is the blended attack \cite{chen2017targeted}, where $x' = \lambda \cdot k + (1 - \lambda) \cdot x$ and $\lambda$ is set to 0.15. The attack target $t$ is set to category 0 for both datasets. When selecting $U$ with FUS, we use VGG-16 \cite{simonyan2014very} as the victim DNN architecture and use SGD with the momentum of 0.9 and the weight decay of 5e-4 as the optimizer. $\alpha$ is set to 0.5 and $N$ is set to 10, if not otherwise specified. The total training duration is 60, and the batch size is 512 for CIFAR-10 and 256 for ImageNet-10. The initial learning rate is set to 0.01 and is dropped by 10 after 30 and 50 epochs.

Once the build of $\mathcal{U}$ is completed, we use it to perform the backdoor injection under two conditions, i.e., the white-box setting and the black-box setting. The first case assumes that the attacker knows in advance the model architecture, the optimizer, and the training hyperparameters used by the user, and therefore can select $\mathcal{U}$ with the same setting. The black-box condition is more realistic, where the attacker is agnostic about the user's configuration. Here, we test $\mathcal{U}$ on four DNN architectures, VGG-13 \cite{simonyan2014very}, VGG-16 \cite{simonyan2014very}, ResNet-18 \cite{he2016deep}, and PreActResNet-18 \cite{he2016identity}, two optimizers, SGD and ADAM \cite{kingma2014adam}, three batch sizes, 128, 256, 512, and four initial learning rates, 0.001, 0.002, 0.01, 0.02, to simulate this situation.

\subsection{Experimental Results}
\begin{figure}[t]
\centering
\includegraphics[width=\oosize]{\fpath/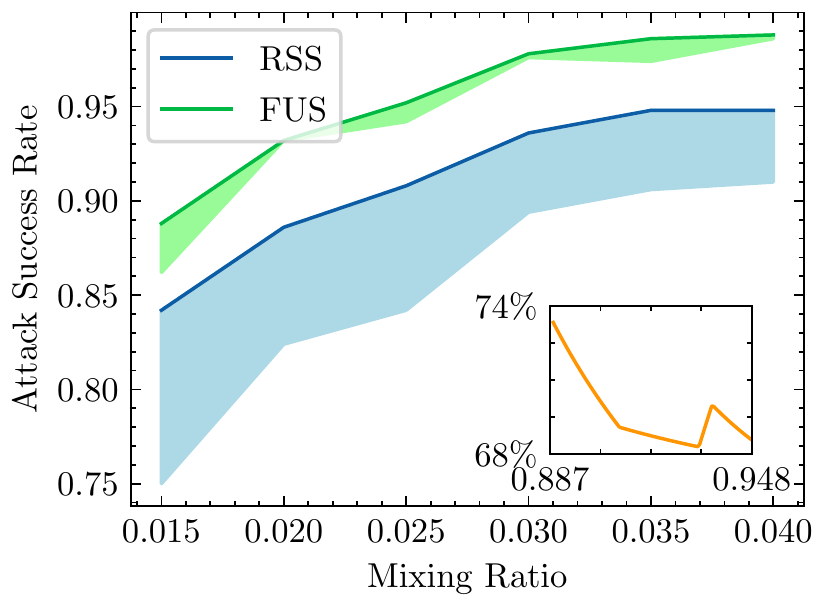}
\caption{White-box result of FUS and RSS on ImageNet-10 and VGG-16. Under the same computing time, the experiment is repeated 3 and 30 times for FUS and RSS, respectively, and the solid lines represent the best runs. The subplot shows the percentage of the FUS-selected sample volume to the RSS-selected sample volume for the same attack strength.} 
\label{fig:result_i10_v16_b00a}
\end{figure}

The white-box results on CIFAR-10 and ImageNet-10 are shown in Figure \ref{fig:result_c10_v16_b00a} and Figure \ref{fig:result_i10_v16_b00a}, where the Random Selected Strategy (RSS) is used as a baseline for comparison. It can be seen that, for different mixing ratios, the attack success rate using the FUS-selected poisoned samples is always better than using the RSS-selected poisoned samples with a large margin. The boosts are about 0.02 to 0.05 for CIFAR-10 and 0.04 to 0.05 for ImageNet-10. To provide the comparison of the data volumes of FUS and RSS for reaching the same attack strength, we calculate the percentage using the linear interpolation and the results are shown in the subplots of Figure \ref{fig:result_c10_v16_b00a} and Figure \ref{fig:result_i10_v16_b00a}. FUS can save 25\% to 53\% of the data volume on CIFAR-10 and 26\% to 32\% of data volume on ImageNet-10 to achieve the same attack success rate as RSS. These results indicate that the proposed method can improve the efficiency of data poisoning in the white-box setting, thereby reducing the number of poisoned samples required. This surely increases the stealthiness of backdoor attacks.

\begin{figure*}[t]
\centering
\subfigure[]{\includegraphics[width=\tosize]{\fpath/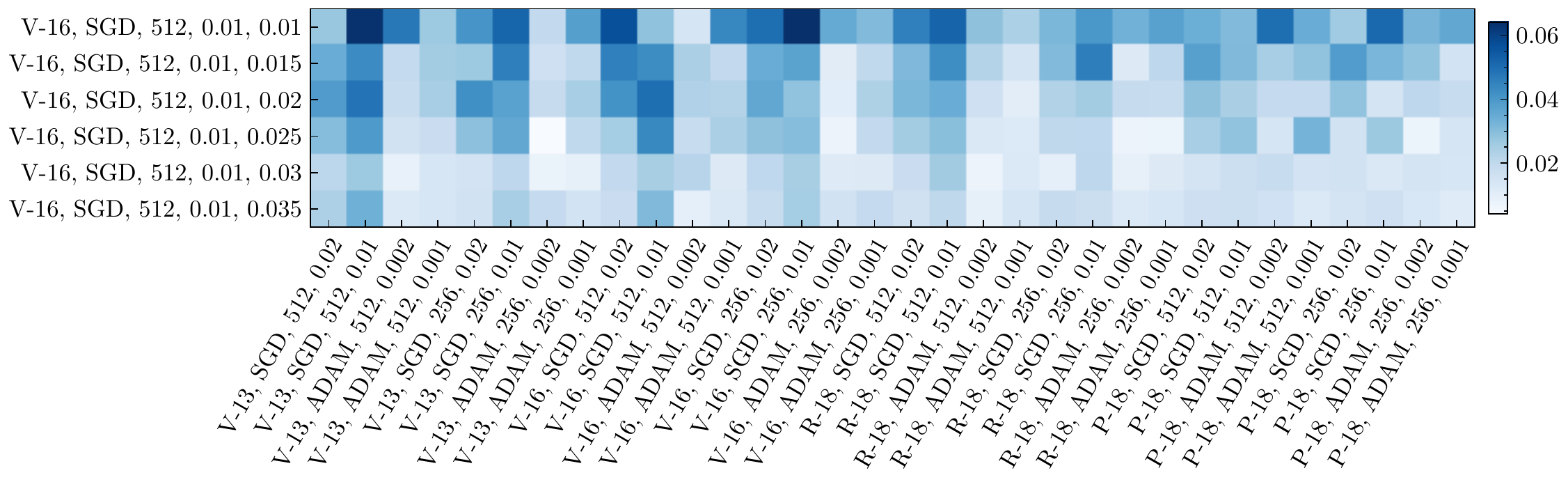}}
\subfigure[]{\includegraphics[width=\tosize]{\fpath/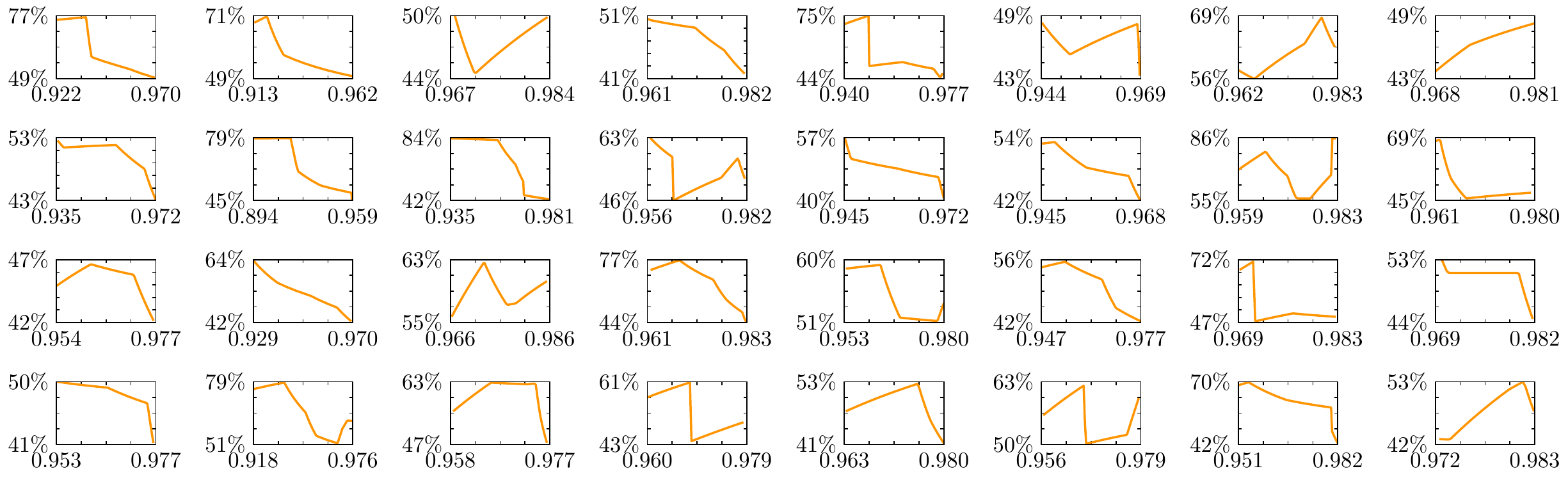}}
\caption{Black-box results of FUS and RSS on CIFAR-10. (a): The difference between the attack success rate using the FUS-selected samples and the rate using the RSS-selected samples, where the vertical axis represents the settings used in the selection of $\mathcal{U}$, including the model, the optimizer, the batch size, the initial learning rate, and the mixing ratio, and the horizontal axis represents the settings used in the backdoor injection, including the model, the optimizer, the batch size, and the initial learning rate. V-13, V-16, R-18, and P-18 denote VGG-13, VGG-16, ResNet-18, and PreActResNet-18, respectively. (b): The percentage of the FUS-selected sample volume to the RSS-selected sample volume in the black-box setting for the same attack strength. The figures correspond to the settings of the horizontal axis of (a) from left to right and from top to bottom, respectively.}
\label{fig:transfer_c10_v16_b00a}
\end{figure*}

In practice, the more common scenario is that the attacker does not know any prior knowledge about the user's configuration. The black-box results on CIFAR-10 are shown in Figure \ref{fig:transfer_c10_v16_b00a}. The results on ImageNet-10 are similar and are presented in the appendix. With multiple black-box settings, using the FUS-selected poisoned samples consistently has a higher success rate than using the RSS-selected samples. The improvements are about 0.01 to 0.06 for CIFAR-10 and 0.01 to 0.08 for ImageNet-10. Likewise, we calculate the percentage of the poisoned sample volumes of the FUS selection to the RSS selection for the same attack intensity. As it can be seen, approximately 14\% to 59\% of the data volume for CIFAR-10 and 9\% to 49\% of the data volume for ImageNet-10 is saved. These results indicate that the FUS-selected samples have good transferability and can be applied in practice, as the method does not require prior knowledge of the model architecture, the optimizer, and the training hyperparameters employed by the user.

\begin{table}[t] 
\centering 
\begin{tabular}{c|c|c|c|c|c|c} 
\toprule
\diagbox{$\alpha$}{$r$} & 0.01              & 0.015             & 0.02              & 0.025             & 0.03              & 0.035            \\
\midrule
0.1                     & 0.909             & 0.946             & 0.960             & 0.967             & 0.978             & 0.980            \\
0.3                     & \underline{0.923} & \underline{0.966} & \underline{0.980} & 0.983             & 0.988             & 0.991            \\
0.5                     & 0.921             & \underline{0.966} & 0.977             & \underline{0.985} & \underline{0.990} & 0.991            \\
0.7                     & 0.913             & 0.955             & 0.976             & 0.981             & 0.987             & \underline{0.992} \\
0.9                     & 0.882             & 0.936             & 0.955             & 0.967             & 0.978             & 0.982            \\
1.0                     & 0.876             & 0.932             & 0.941             & 0.956             & 0.963             & 0.970            \\
\bottomrule
\end{tabular}
\caption{Attack success rate with different $\alpha$ on CIFAR-10 and VGG-16, where the underlines highlight the best values for each column.}
\label{tab:alpha_c10_v16_b00a}
\end{table}

\begin{figure}[t]
\centering
\includegraphics[width=\oosize]{\fpath/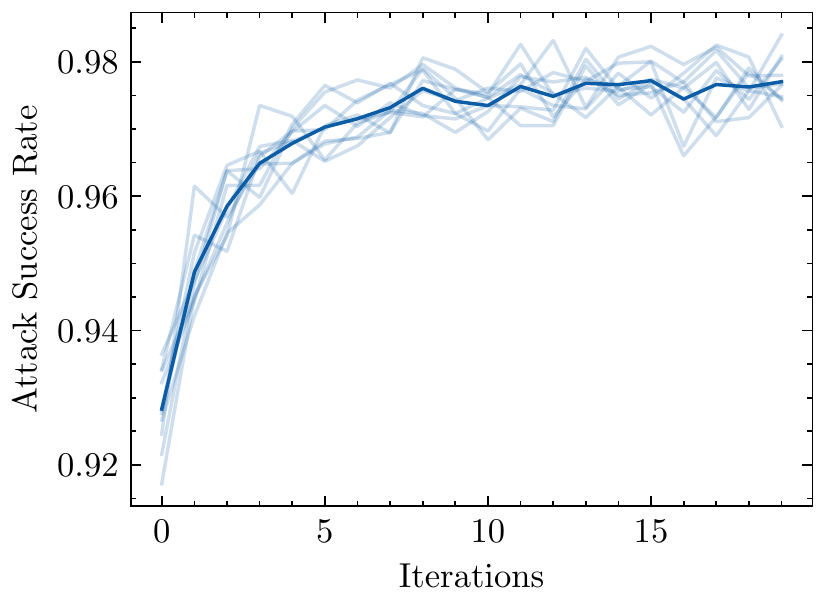}
\caption{Attack success rate with $N = 20$ on CIFAR-10 and VGG-16 when $r = 0.02$. The solid line represents the average of 10 runs.} 
\label{fig:iteration_i10_v16_b00a}
\end{figure}

\subsection{Ablation Studies}
We conduct ablation studies of the hyperparameters in FUS, i.e., $\alpha$ and $N$, and the results are shown in Table \ref{tab:alpha_c10_v16_b00a} and Figure \ref{fig:iteration_i10_v16_b00a}, respectively. $\alpha$ represents the proportion of the sample pool that is filtered out each time and has a relatively large effect on FUS. $\alpha$ that is either too small or too large leads to a degradation of FUS's performance, with the former causing a slower update of the sample pool and the latter causing a failure of the algorithm to converge. Numerically, FUS performs best with $\alpha$ set to 0.3 or 0.5. $N$ represents the number of iterations of FUS. When $N = 0$, FUS degenerates to RSS, and when $N = 1$, FUS is equivalent to greedily selecting the high-contribution poisoned samples. As can be seen from Figure \ref{fig:iteration_i10_v16_b00a}, the success rate of the attack grows gradually as the iteration proceeds. This indicates that the previously mentioned local selection problem does exist, and that our method can alleviate it to some extent. Considering the time consumption and the slowing down of the growth rate when $N$ is greater than 10, we set $N$ to 10 in this paper.

\subsection{Attribution Study}
\begin{figure}[h]
\centering
\subfigure[]{\includegraphics[width=7.8cm]{\fpath/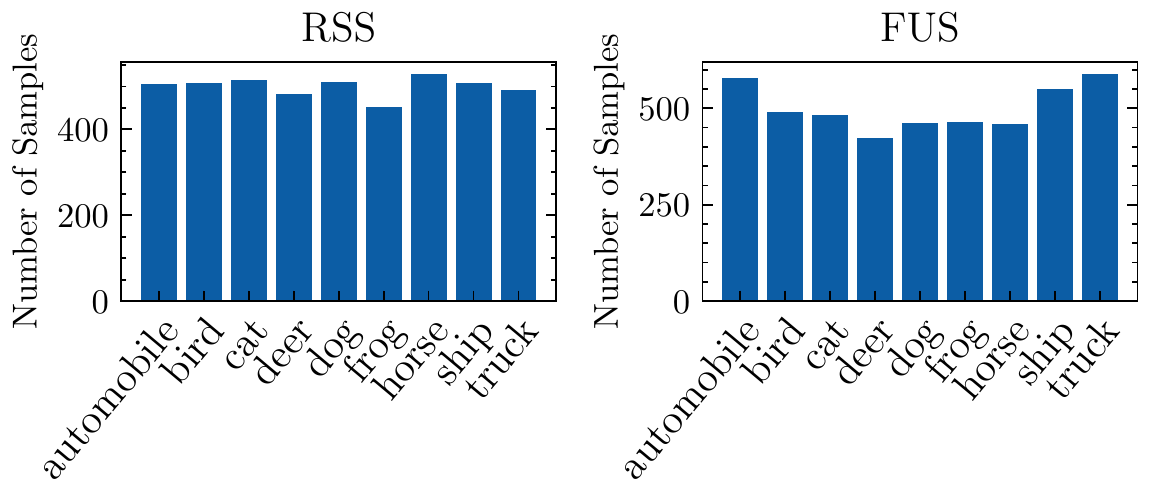}}
\subfigure[]{\includegraphics[width=7.8cm]{\fpath/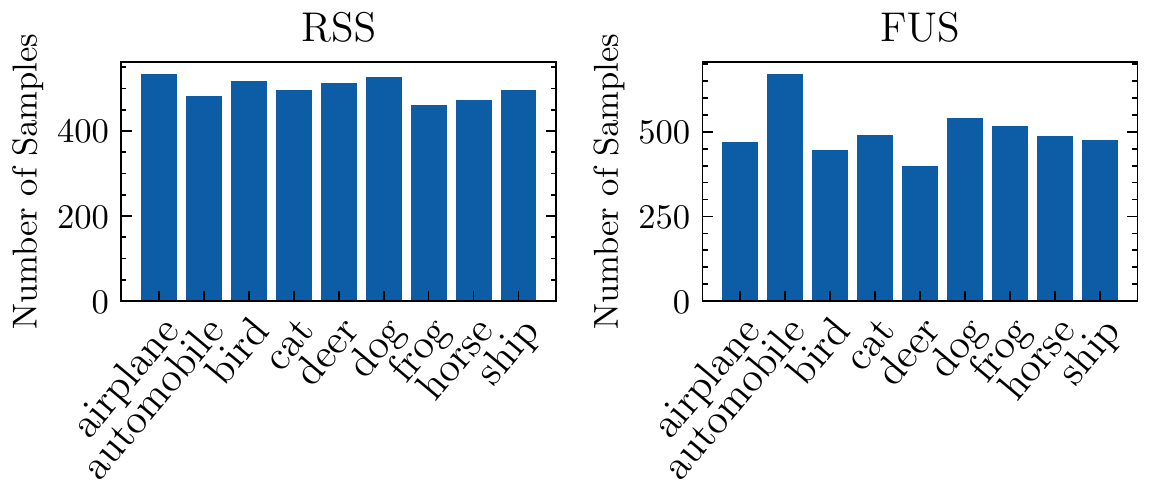}}
\caption{Original category statistics for the FUS-selected samples and the RSS-selected samples with different $t$ on CIFAR-10 and VGG-16 when $r = 0.03$. (a): $t$ is set to category 0, ``airplane''. (b): $t$ is set to category 9, ``truck''.}
\label{fig:hist_iteration_c10_v16_b0a}
\end{figure}

In this part, we want to know what makes these adversaries selected by FUS poison efficiently. The first reason we consider is that FUS may select more samples from the categories associated with the attack target $t$. Therefore, we count the original classes for the FUS-selected samples and the RSS-selected samples with different $t$, as shown in Figure \ref{fig:hist_iteration_c10_v16_b0a}. The results seem to show that our assumption is correct. For example, when $t$ is set to 9, ``truck'', the most original category of the FSS-selected poisoned samples is ``automobile''. 

Naturally, the next question is whether the adversaries sampled randomly based on the same class distribution as the FUS-selected samples would yield the same attack performance, too. We experiment and the result is shown in Table \ref{tab:dist_c10_v16_b0a}. The attack success rates of the poisoned samples selected using RSS are similar in the two distributions. This indicates that the fundamental reason why the FUS-selected samples work well is not because of the class distribution, but rather the samples themselves. The class distribution is a symptom, not a cause.

\begin{table}[t]
\centering 
\begin{tabular}{c|c|c|c|c} 
\toprule
\multirow{2}*{Target} & \multicolumn{2}{c|}{RSS with U} & \multicolumn{2}{c}{RSS with S} \\
                      & mean  & max                     & mean & max                      \\
\midrule
``airplane''          & 0.954 & 0.963                   & 0.955 & 0.963                   \\
``cat''               & 0.955 & 0.966                   & 0.955 & 0.965                   \\
``dog''               & 0.954 & 0.962                   & 0.955 & 0.962                   \\
``truck''             & 0.957 & 0.964                   & 0.958 & 0.968                   \\
\bottomrule
\end{tabular}
\caption{Attack success rate of RSS on CIFAR-10 and VGG-16 with the Uniform distribution (U) and RSS with the same class distribution (S) as the FUS-selected samples. Each experiment is repeated 10 times.}
\label{tab:dist_c10_v16_b0a}
\end{table}

\section{Conclusion}
The selection of poisoned samples is important for the efficiency of poisoning in backdoor attacks. Existing methods use the random selection strategy, which ignores the fact that each adversary contributes differently to the backdoor injection. It reduces the efficiency and further raises the probability of the attack being detected. In this paper, we formulate the selection as an optimization problem and propose a strategy named FUS to solve it. Experiments on CIFAR-10 and ImageNet-10 are conducted to test our method. In the white-box setting, FUS can save about 25\% to 53\% of the poisoned data volume to reach the same attack strength as the random selection strategy. In the black-box setting, the value is about 9\% to 59\%. These results indicate that FUS can increase the efficiency of poisoning data and thus the stealthiness of the attack.

\section*{Acknowledgments}
The work was supported in part by the National Natural Science Foundation of China under Grands U19B2044 and 61836011.

\bibliographystyle{named}
\bibliography{./references.bib}
\end{document}